\documentclass[runningheads]{llncs}
\usepackage{graphicx}
\usepackage{amsmath,amssymb} 
\usepackage{color}

\usepackage{mathtools}

\usepackage{enumitem}

\usepackage{subfigure}

\newcommand{\etal}{\textit{et al}.}

\def \x{\mathbf{x}}
\def \f{\mathbf{f}}
\def \y{\mathbf{y}}

\DeclareMathOperator*{\argmax}{arg\,max}

\begin{document}
\pagestyle{headings}
\mainmatter

\title{Real-Time MDNet} 

\titlerunning{Real-Time MDNet}

\authorrunning{I. Jung, J. Son, M. Baek, and B. Han}

\author{Ilchae Jung$^{1}$, Jeany Son$^{1}$, Mooyeol Baek$^{1}$, and Bohyung Han$^{2}$ }


\institute{$^{1}$Dept. of CSE, POSTECH, Korea\\
	$^{2}$Dept. of ECE \& ASRI, Seoul National University, Korea\\
	\email{ \{chey0313,jeany,mooyeol\}@postech.ac.kr}
	\email{ bhhan@snu.ac.kr}
}

\maketitle

\begin{abstract}
We present a fast and accurate visual tracking algorithm based on the multi-domain convolutional neural network (MDNet).
The proposed approach accelerates feature extraction procedure and learns more discriminative models for instance classification; it enhances representation quality of target and background by maintaining a high resolution feature map with a large receptive field per activation.
We also introduce a novel loss term to differentiate foreground instances across multiple domains and learn a more discriminative embedding of target objects with similar semantics.
The proposed techniques are integrated into the pipeline of a well known CNN-based visual tracking algorithm, MDNet.
We accomplish approximately 25 times speed-up with almost identical accuracy compared to MDNet.
Our algorithm is evaluated in multiple popular tracking benchmark datasets including OTB2015, UAV123, and TempleColor, and outperforms the state-of-the-art real-time tracking methods consistently even without dataset-specific parameter tuning.


\end{abstract}



\section{Introduction}
\label{sec:introduction}

Recently, Convolutional Neural Networks (CNNs) are very effective in visual tracking~\cite{nam2016learning,tcnn,eco,ccot,adnet,sanet,fcn,cnnsvm,tsn,branchout}, but, unfortunately, highly accurate tracking algorithms based on CNNs are often too slow for practical systems.
There are only a few methods~\cite{bacf,ptav,east} that achieve two potentially conflicting goals---accuracy and speed---at the same time.

MDNet~\cite{nam2016learning} is a popular CNN-based tracking algorithm with state-of-the-art accuracy.
This algorithm is inspired by an object detection network, R-CNN~\cite{rcnn}; it samples candidate regions, which are passed through a CNN pretrained on a large-scale dataset and fine-tuned at the first frame in a test video.
Since every candidate is processed independently, MDNet suffers from high computational complexity in terms of time and space.
In addition, while its multi-domain learning framework concentrates on saliency of target against background in each domain, it is not optimized to distinguish potential target instances across multiple domains. 
Consequently, a learned model by MDNet is not effective to discriminatively represent unseen target objects with similar semantics in test sequences.

A straightforward way to avoid redundant observations and accelerate inference is to perform RoIPooling from a feature map~\cite{fastrcnn}, but na{\"i}ve implementations result in poor localization due to coarse quantization of the feature map.
To alleviate such harsh quantization for RoI pooling, \cite{maskrcnn} proposes RoI alignment (RoIAlign) via bilinear interpolation.
However, it may also lose useful localization cues within target if the size of RoI is large.
On the other hand, since most CNNs are pretrained for image classification tasks, the networks are competitive to predict semantic image labels but insensitive to tell differences between object instances in low- or mid-level representations.
A direct application of such CNNs to visual tracking often yields degradation of accuracy since the embedding generated by pretrained CNNs for image classification task is not effective to differentiate two objects in the same category.

To tackle such critical limitations, we propose a novel real-time visual tracking algorithm based on MDNet by making the following contributions.
First, we employ an RoIAlign layer to extract object representations from a preceding fully convolutional feature map.
To maintain object representation capacity, the network architecture is updated to construct a high resolution feature map and enlarge the receptive field of each activation.
The former is helpful to represent candidate objects more precisely, and the latter is to learn rich semantic information of target.
Second, we introduce an instance embedding loss in pretraining stage and aggregate to the existing binary foreground/background classification loss employed in the original MDNet.
The new loss function plays an important role to embed observed target instances apart from each other in a latent space.
It enables us to learn more discriminative representations of unseen objects even in the case that they have identical class labels or similar semantics.

Our main contributions are summarized as follows:
\begin{itemize}[label=$\bullet$]
  \item We propose a real-time tracking algorithm inspired by MDNet and Fast R-CNN, where an improved RoIAlign technique is employed to extract more accurate representations of targets and candidates from a feature map and improve target localization.
  \item We learn shared representations using a multi-task loss in a similar way to MDNet, but learn an embedding space to discriminate object instances with similar semantics across multiple domains more effectively.
  \item The proposed algorithm demonstrates outstanding performance in multiple benchmark datasets without dataset-specific parameter tuning.
Our tracker runs real-time with 25 times speed-up compared to MDNet while maintaining almost identical accuracy.
\end{itemize}

The rest of the paper is organized as follows.
We first discuss related work in Section~\ref{sec:related}.
Section~\ref{sec:proposed} discusses our main contribution for target representation via the improved RoIAlign and the instance embedding loss.
We present overall tracking algorithm in Section~\ref{sec:tracking}, and provide experimental results in Section~\ref{sec:experiments}.
We conclude this paper in Section~\ref{sec:conclusion}.


\section{Related Work}
\label{sec:related}

\subsection{Visual Tracking Algorithms}

CNN-based visual tracking algorithms typically formulate object tracking as discriminative object detection problems.
Some methods~\cite{nam2016learning,tcnn,adnet,sanet,tsn,branchout} draw a set of samples corresponding to candidate regions and compute their likelihoods independently using CNNs.
Recent techniques based on discriminative correlation filters boost accuracy significantly by incorporating representations from deep neural networks~\cite{hcf,crest,eco,ccot,mcpf}.
Although various tracking algorithms based on CNNs are successful in terms of accuracy, they often suffer from high computational cost mainly due to critical time consuming components within the methods including feature computation of multiple samples, backpropagation for model updates, feature extraction from deep networks, etc.
While some CNN-based techniques~\cite{goturn,siamfc,east} for visual tracking run real-time by employing offline representation learning without online model updates, their accuracy is not competitive compared to the state-of-the-art methods.

There are only a few real-time trackers~\cite{eco,bacf,ptav} that present competitive accuracy.
Galoogahi~\etal~\cite{bacf} incorporate background region to learn more discriminative correlation filters using hand-crafted features.
Fan~\etal~\cite{ptav} design a robust tracking algorithm through interactions between a tracker and a verifier.
Tracker estimates target states based on the observation using hand-crafted features efficiently while verifier double-checks the estimation using the features from deep neural networks.
Danelljan \etal~\cite{eco} propose a discriminative correlation filter for efficient tracking by integrating multi-resolution deep features. 
Since its implementation with deep representations is computationally expensive, they also introduce a high-speed tracking algorithm with competitive accuracy based on hand-crafted features.
Note that most real-time trackers with competitive accuracy rely on hand-crafted features or limited use of deep representations.
Contrary to such real-time tracking methods, our algorithm has a simpler inference pipeline within a pure deep neural network framework.

\subsection{Representation Learning for Visual Tracking}

MDNet~\cite{nam2016learning} pretrains class-agnostic representations appropriate for visual tracking task by fine-tuning a CNN trained originally for image classification.
It deals with label conflict issue across videos by employing multi-domain learning, and achieves the state-of-the-art performance in multiple datasets.
Since the great success of MDNet~\cite{nam2016learning}, there have been several attempts to learn representations for visual tracking~\cite{siamfc,CFNET,cfcf} using deep neural networks.
Bertinetto~\etal~\cite{siamfc} learn to maximize correlation scores between the same objects appearing in different frames.
Valmadre~\etal~\cite{CFNET} regress response maps between target objects and input images to maximize the score at the ground-truth target location. 
Similarly, Gundogdu~\etal~\cite{cfcf} train deep features to minimize difference between the response map from a tracker based on correlation filters and the ground-truth map that has a peaky maximum value at target location.

All the efforts discussed above focus on how to make target objects salient against backgrounds. 
While this strategy is effective to separate target from background, it is still challenging to discrimninate between object instances with similar semantics.
Therefore, our algorithm encourages our network to achieve the two objectives jointly by proposing a novel loss function with two terms.

\subsection{Feature Extraction in Object Detection}
Although R-CNN~\cite{rcnn} is successful in object detection, it has significant overhead to extract features from individual regions for inference.
Fast R-CNN~\cite{fastrcnn} reduces its computational cost for feature extraction using RoIPooling, which computes fixed-size feature vectors by applying max pooling to the specific regions in a feature map.
While the benefit in terms of computational cost is impressive, RoIPooling is not effective to localize targets because it relies on a coarse feature map.
To alleviate this limitation, mask R-CNN~\cite{maskrcnn} introduces a new feature extraction technique, RoI alignment (RoIAlign), which approximates features via bilinear interpolation for better object localization.
Our work proposes a modified network architecture for an adaptive RoIAlign to extract robust features corresponding to region proposals.


\section{Efficient Feature Extraction and Discriminative Feature Learning}
\label{sec:proposed}

This section describes our CNN architecture with an improved RoiAlign layer, which accelerates feature extraction while maintaining quality of representations.
We also discuss a novel multi-domain learning approach with discriminative instance embedding of foreground objects.

\subsection{Network Architecture}

\begin{figure}[t]
\centering
\includegraphics[width=1\linewidth]{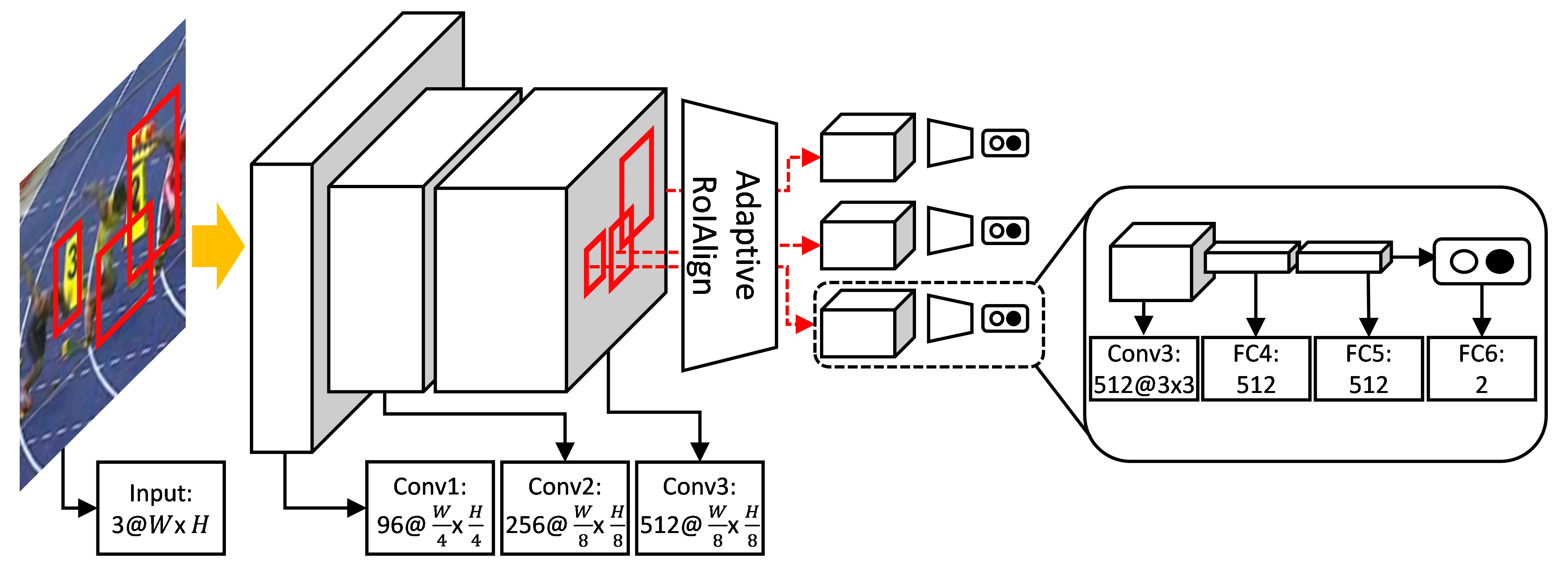}
\caption{Network architecture of the proposed tracking algorithm. The network is composed of three convolutional layers for extracting a shared feature map, adaptive RoIAlign layer for extracting a specific feature using regions of interest (RoIs), and three fully connected layers for binary classification. The number of channels and the size of each feature map are shown with the name of each layer.}
\label{fig:structure}
\end{figure}

Fig.~\ref{fig:structure} illustrates architecture of our model.
The proposed network consists of fully convolutional layers (\texttt{conv1-3}) for constructing a shared feature map, an adaptive RoIAlign layer for extracting feature of each RoI, and three fully connected layers (\texttt{fc4-6}) for binary classification.
Given a whole image with a set of proposal 
bounding boxes as an input, the network computes a shared feature map of the input image through a single forward pass.
A CNN feature corresponding to each RoI is extracted from the shared feature map using an adaptive RoIAlign operation.
Through this feature computation strategy, we reduce computational complexity significantly while improving quality of features.

The extracted feature representation from each RoI is fed to two fully connected layers for classification between target and background.
We create multiple branches of domain specific layers (\texttt{fc$6^1$-fc$6^D$}) for multi-domain learning, and learn a discriminative instance embedding.  
During online tracking, a set of the domain-specific fully connected layers are replaced by a single binary classification layer with softmax cross-entropy loss, which will be fine-tuned using the examples from an initial frame.

\subsection{Improved RoIAlign for Visual Tracking}
\label{subsec:RoIalign}

\begin{figure}[ht]
\centering
\subfigure[Original VGG-M network]{
\includegraphics[width=0.5\linewidth]{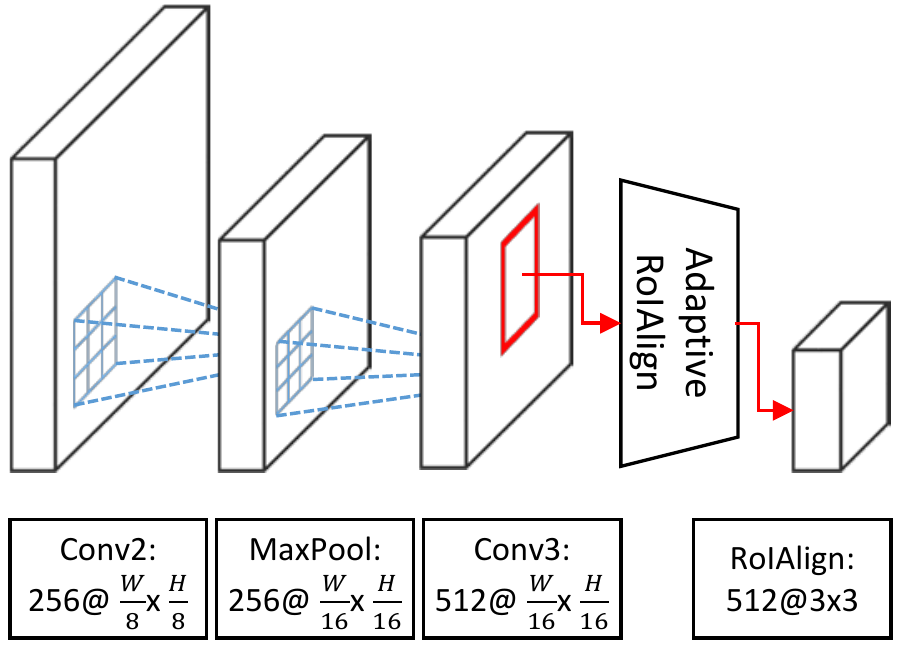}
}
\subfigure[Network for dense feature map]{
\includegraphics[width=0.415\linewidth]{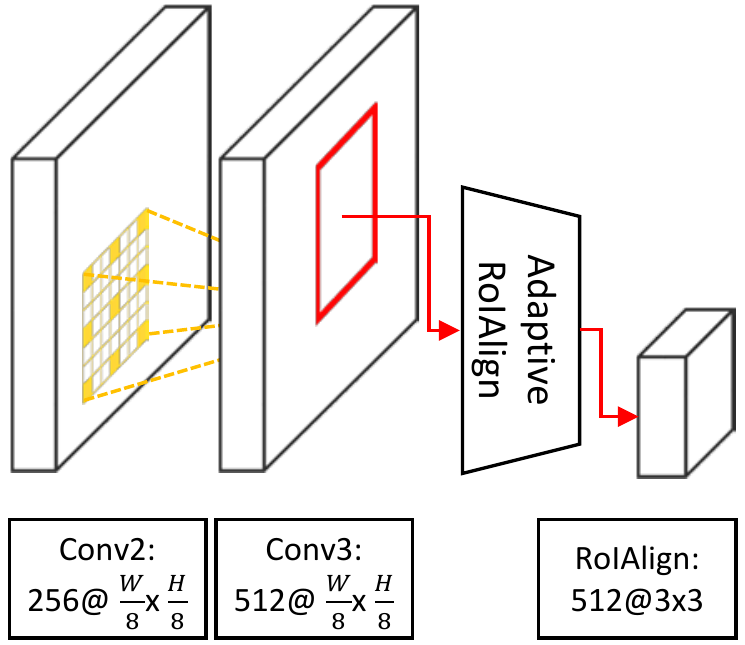}
}
\caption{Network architecture of the part of our fully convolutional network for extracting a shared feature map. Max pooling layer is removed after conv2 layer in original VGG-M network, and dilated convolution with rate $r=3$ is applied for extracting a dense feature map with a higher spatial resolution.}
\label{fig:fcnetwork}
\end{figure}

Our network has an RoIAlign layer to obtain object representations from a fully convolutional feature map constructed from a whole image.
However, features extracted by RoIAlign are inherently coarse compared to the ones from individual proposal bounding boxes.
To improve quality of representations of RoIs, we need to construct a feature map with high resolution and rich semantic information.
These requirements can be addressed by computing a denser fully convolutional feature map and enlarging the receptive field of each activation.
To these ends, we remove a max pooling layer followed by \texttt{conv2} layer in VGG-M network~\cite{vggm} and perform dilated convolutions~\cite{deeplab} in \texttt{conv3} layer with rate $r=3$.
This strategy results in a twice larger feature map than the output of \texttt{conv3} layer in the original VGG-M network.
It allows to extract high resolution features and improve quality of representation.
Fig.~\ref{fig:fcnetwork} compares our network for dense feature map computation with the original VGG-M network.

Our adaptive RoIAlign layer computes more reliable features, especially for large objects, using a modified bilinear interpolation.
Since ordinary RoIAlign only utilizes nearby grid points on the feature map to compute the interpolated value, it may lose useful information if the interval of the sampled points for RoI is larger than the one of the feature map grid.
To handle this issue, we adjust the interval of the grid points from the shared dense feature map adaptively. 
In specific, the bandwidth of the bilinear interpolation is determined by the size of RoIs; it is proportional to $[\frac{w}{w'}]$, where $w$ and $w'$ denote the width of RoI after the \texttt{conv3} layer and the width of RoI's output feature in the RoIAlign layer, respectively, and $[\cdot]$ is a rounding operator.

The technique integrating a network to employ a dense feature map and the adaptive RoIAlign  is referred to as improved RoIAlign.
Our adaptive RoIAlign layer produces a $7 \times 7$ feature map, and a max pooling layer is applied after the layer to produce a $3\times3$ feature map.
Although the improved RoIAlign makes minor changes, it improves performance of our tracking algorithm significantly in practice.
This is partly because, on the contrary to object detection, tracking errors originated from subtle differences in target representations are propagated over time and create large errors to make trackers fail eventually.

\subsection{Pretraining for Discriminative Instance Embedding}
\label{subsec:learning}

\begin{figure}[t]
\centering
\includegraphics[width=0.9\linewidth]{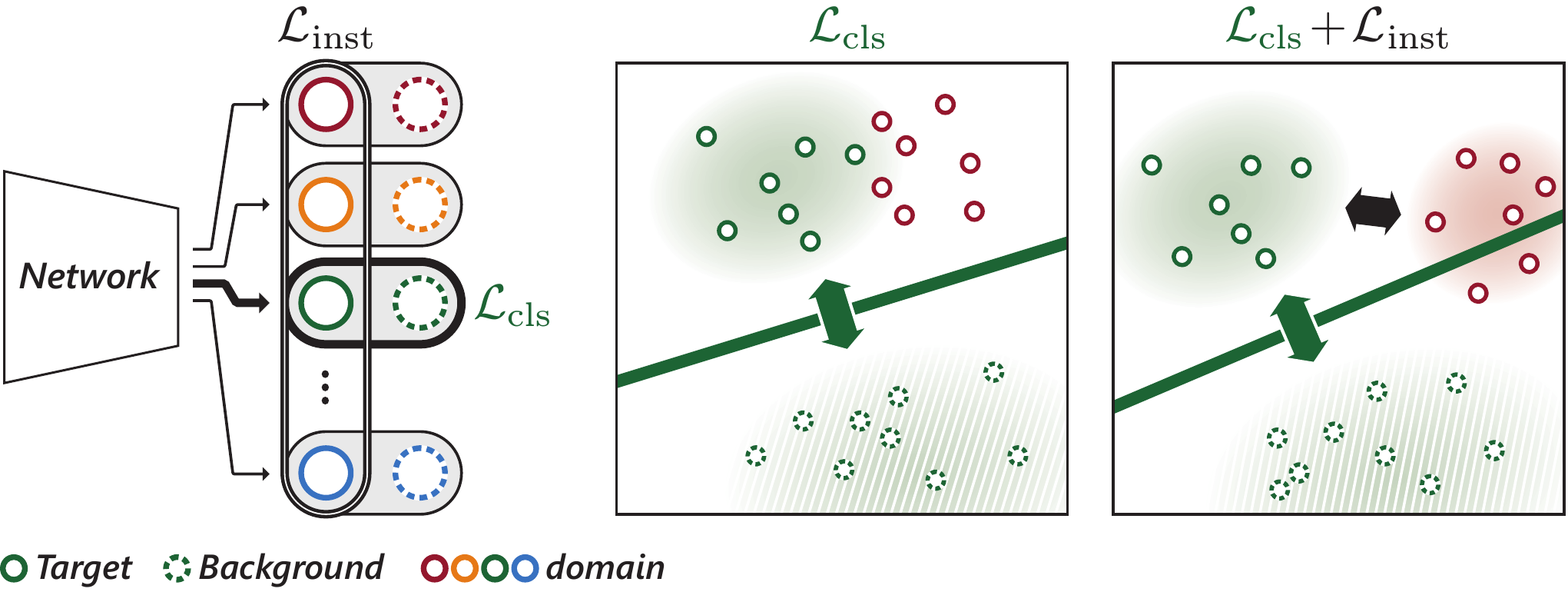}
\caption{Multi-task learning for binary classification of the target object and instance embedding across multiple domains. 
The binary classification loss is designed to distinguish the target and background, while the instance embedding loss separates target instances. Note that a minibatch in each iteration for training is constructed by sampling from a single domain.}
\label{fig:jointlearning}
\end{figure}

The goal of our learning algorithm is to train a discriminative feature embedding applicable to multiple domains.
MDNet has the separate shared and domain-specific layers to learn the representations distinguishing between target and background.
In addition to this objective, we propose a new loss term, referred to as an instance embedding loss, which enforces target objects in different domains to be embedded far from each other in a shared feature space and enables to learn discriminative representations of the unseen target objects in new test sequences.
In other words, MDNet only attempts to discriminate target and background in individual domains, and may not be powerful to discriminate foreground objects in different domains, especially when the foreground objects belong to the same semantic class or have similar appearances.
This is partly because the original CNNs are trained for image classification.
To handle this issue, our algorithm incorporates an additional constraint, which embeds foreground objects from multiple videos to be apart from each other.

Given an input image $\x^d$ in domain $d$ and a bounding box $R$, the output score of the network, denoted by $\f^d$, is constructed by concatenating the activations from the last fully connected layers (\texttt{fc$6^1$-fc$6^D$}) as
\begin{align}
    \f^d = [\phi^1(\x^d;R), \phi^2(\x^d;R), \dots, \phi^D(\x^d;R)]\in \mathbb{R}^{2\times D},
\end{align}
where $\phi^d(\cdot;\cdot)$ is a 2D binary classification score from the last fully connected layer \texttt{fc$6^d$ in domain $d$,} and $D$ is the number of domains in a training dataset.
The output feature is given to a softmax function for binary classification, which determines whether a bounding box $R$ is a target or a background patch in domain $d$.
Additionally, the output feature is passed through another softmax operator for discriminating instances in multiple domains.
The two softmax functions are given by
\begin{align}
    [\sigma_{\text{cls}}(\f^d)]_{ij} = \frac{\exp({f_{ij}^d})}{\sum_{k=1}^2 \exp({f_{kj}^d})} ~~\text{and}~~
    [\sigma_{\text{inst}}(\f^d)]_{ij} = \frac{\exp({f_{ij}^d})}{\sum_{k=1}^D \exp({f_{ik}^d})},
\end{align}
where $\sigma_{\text{cls}}(\cdot)$ compares scores of target and background in each domain whereas $\sigma_{\text{inst}}(\cdot)$ compares the positive scores of the objects across all domains.

Our network minimizes a multi-task loss $\mathcal{L}$ on the two softmax operators, which is given by
\begin{align}
    \mathcal{L} = \mathcal{L}_\text{cls}+ \alpha \cdot \mathcal{L}_\text{inst},
\label{eq:loss}
\end{align}
where $\mathcal{L}_\text{cls}$ and $\mathcal{L}_\text{inst}$ are loss terms for binary classification and discriminative instance embedding, respectively, and $\alpha$ is a hyper-parameter that controls balance between the two loss terms.
Following MDNet, we handle a single domain in each iteration; the network is updated based on a minibatch collected from the $(k~\text{mod}~ D)^\text{th}$ domain only in the $k^\text{th}$ iteration.

The binary classification loss with domain $\hat d(k) = (k~\text{mod}~ D)$ in the $k^\text{th}$ iteration is given by 
\begin{align}
\mathcal{L}_\text{cls} = -\frac{1}{N}\sum_{i=1}^N \sum_{c=1}^2 [\y_{i}]_{c\hat d(k)} \cdot \log\left(\left[\sigma_\text{cls}\left(\f^{\hat d(k)}_i\right)\right]_{c \hat d(k)}\right),
\end{align}
where $\y_{i} \in \{0,1\}^{2\times D}$ is a one-hot encoding of a ground-truth label; its element $[\y_i]_{cd}$ is 1 if a bounding box $R_i$ in domain $d$ corresponds to class c, otherwise 0.
Also, the loss for discriminative instance embedding is given by
\begin{align}
\mathcal{L}_\text{inst} = -\frac{1}{N}\sum_{i=1}^N \sum_{d=1}^D [\y_{i}]_{+d} \cdot \log\left(\left[\sigma_\text{inst}\left(\f^d_i\right)\right]_{+d}\right).
\label{eq:instance_loss}
\end{align}
Note that the instance embedding loss is applied only to positive examples using the positive channel denoted by $+$ in Eq.~\eqref{eq:instance_loss}.
As a result of the proposed loss, the positive scores of target objects in current domain become larger while their scores in other domains get smaller.
It leads to a distinctive feature embedding of target instances and makes it effective to distinguish similar objects potentially appearing in new testing domains.

Fig.~\ref{fig:jointlearning} illustrates impact of the multi-task learning for discriminative feature embedding of target instances across multiple domains.

\section{Online Tracking Algorithm}
\label{sec:tracking}
We discuss the detailed procedure of our tracking algorithm including implementation details.
The pipeline of our tracking algorithm is almost identical to MDNet~\cite{nam2016learning}.

\subsection{Main Loop of Tracking}
Once pretraining is completed, we replace multiple branches of domain-specific layers (\texttt{fc$6^1$-fc$6^D$}) with a single branch for each test sequence.
Given the first frame with ground-truth of target location, we fine-tune fully connected layers (\texttt{fc4-6}) and customize the network to a test sequence.
For the rest of frames, we update the fully connected layers in an online manner while convolutional layers are fixed.
Given an input frame at time $t$, a set of samples, denoted by $\{\x_t^i\}_{i=1 \dots N}$, are drawn from a Gaussian distribution centered at the target state of the previous frame, the optimal target state is given by
\begin{align}
    \x_t^* = \argmax_{\x_t^i}f^{+}(\x_t^i),
\end{align}
where $f^{+}(\x_t^i)$ indicates the positive score of the $i^{\text{th}}$ sample drawn from the current frame at time step $t$.
Note that tracking is performed in a three dimensional state space for translation and scale change.

We also train a bounding box regressor to improve target localization accuracy motivated by the success in \cite{nam2016learning}. 
Using a set of extracted features from RoIs from the first frame of a video, $\mathcal{F}_{i}^\text{RoI}$, we train a simple linear regressor in the same way to \cite{rcnn,dpm}. 
We apply the learned bounding box regressor from the second frame and adjust the estimated target regions if the estimated target state is sufficiently reliable, $f^{+}(\x_t^*) > 0.5$.

\subsection{Online Model Updates}
 
We perform two complementary update strategies as in MDNet~\cite{nam2016learning}: long-term and short-term updates to maintain robustness and adaptiveness, respectively. 
Long-term updates are regularly applied using the samples collected for a long period of time, while short-term updates are triggered whenever the score of the estimated target is below a threshold and the result is unreliable. 

A minibatch is composed of 128 examples---32 positive and 96 negative samples, for which we employ hard minibatch mining in each iteration of online learning procedure.
The hard negative examples are identified by testing 1024 negative examples and selecting the ones with top 96 positive scores.

\subsection{Implementation Details}

\subsubsection{Network initialization and input management} 
The weights of three convolutional layers are transferred from the corresponding parts in VGG-M network~\cite{vggm} pretrained on ImageNet~\cite{ILSVRC15} while fully connected layers are initialized randomly.
An input image is resized to make the size of target object fit to $107\times 107$, and cropped to the smallest rectangle enclosing all sample RoIs.
The receptive field size of a single unit in
the last convolutional layer is equal to $75\times 75$. 

\subsubsection{Offline pretraining} 
For each iteration of offline pretraining, we construct a minibatch with samples collected from a single domain. 
We first sample 8 frames randomly in the selected domain, and draw 32 positive and 96 negative examples from each frame, which results in 256 positive and 768 negative data altogether in a minibatch. 
The positive bounding boxes have overlap larger than 0.7 with ground-truths in terms of Intersection over Union (IoU) measure while the negative samples have less than 0.5 IoUs. 
Instead of backpropagating gradients in each iteration, we accumulate the gradients from backward passes in multiple iterations; the network is updated at every 50 iteration in our experiments.
We train our models on ImageNet-Vid~\cite{ILSVRC15}, which is a large-scale video dataset for object detection. 
Since this dataset contains a lot of video sequences, almost 4500 videos, we randomly choose 100 videos for an instance embedding loss in each iteration.
Hyper-parameter $\alpha$ in Eq.~\eqref{eq:loss} is set to 0.1.

\subsubsection{Online training} 
Since pretraining stage aims to learn generic representation for visual tracking, we have to fine-tune the pretrained network at the first frame of each testing video.
We draw 500 positive and 5000 negative samples based on the same IoU criteria with the pretraining stage. 
From the second frame, the training data for online updates are collected after tracking is completed in each frame. 
The tracker gather 50 positive and 200 negative examples that have larger than 0.7 IoU and less than 0.3 IoU with the estimated target location, respectively.
Instead of storing the original image patches, our algorithm keep their feature representations to save time and memory by avoiding redundant computation.
Long-term updates are executed every 10 frame.

\subsubsection{Optimization} 
Our network is trained by a Stochastic Gradient Descent (SGD) method. 
For offline representation learning, we train the network for 1000 epochs with learning rate 0.0001 while it is trained for 50 iterations at the first frame of a test video.
For online updates, the number of iterations for fine-tuning is 15 and the learning rate is set to 0.0003. 
The learning rate for \texttt{fc6} is 10 times bigger than others (\texttt{fc4-5}) to facilitate convergence in practice. 
The weight decay and momentum are fixed to 0.0005 and 0.9, respectively. 
Our algorithm is implemented in PyTorch with 3.60 GHz Intel Core I7-6850K and NVIDIA Titan Xp Pascal GPU.


\section{Experiments}
\label{sec:experiments}

This section presents our results on multiple benchmark datasets with comparisons to the state-of-the-art tracking algorithms, and analyzes performance of our tracker by ablation studies.

\subsection{Evaluation Methodology}
We evaluate our tracker, denoted by real-time MDNet or RT-MDNet, on three standard datasets including OTB2015~\cite{otb2015}, UAV123~\cite{uav123} and TempleColor~\cite{templecolor}.
For comparison, we employ several state-of-the-art trackers including ECO~\cite{eco}, MDNet~\cite{nam2016learning}, MDNet+IEL, SRDCF~\cite{srdcf}, C-COT~\cite{ccot}, and top performing real-time trackers, ECO-HC~\cite{eco}, BACF~\cite{bacf}, PTAV~\cite{ptav}, CFNet~\cite{CFNET}, SiamFC~\cite{siamfc} and DSST~\cite{dsst}.
ECO-HC is a real-time variant of ECO based on hand-crafted features, HOG and color names, while MDNet+IEL is a version of MDNet with the instance embedding loss.
Both MDNet and MDNet+IEL are pretrained on IMAGENET-VID.

We follow the evaluation protocol presented in a standard benchmark~\cite{otb2015}, where performance of trackers is evaluated based on two criteria---bounding box overlap ratio and center location error---and is visualized by success and precision plots.
The two plots are generated by computing ratios of successfully tracked frames at a set of different thresholds in the two metrics. 
The Area Under Curve (AUC) scores of individual trackers are used to rank the trackers in the success plot.
In the precision plots, the ranks of trackers are determined by the accuracy at 20 pixel threshold. 
In both plots, real-time trackers are represented with solid lines while the rests are denoted by dashed lines.
Note that the parameters of our algorithm are fixed throughout the experiment; we use the same parameters for all three tested datasets while others may have the different parameter setting for each dataset.

\subsection{Evaluation on OTB2015}

\begin{figure}[t!]
\centering
    \includegraphics[width=0.49\linewidth]{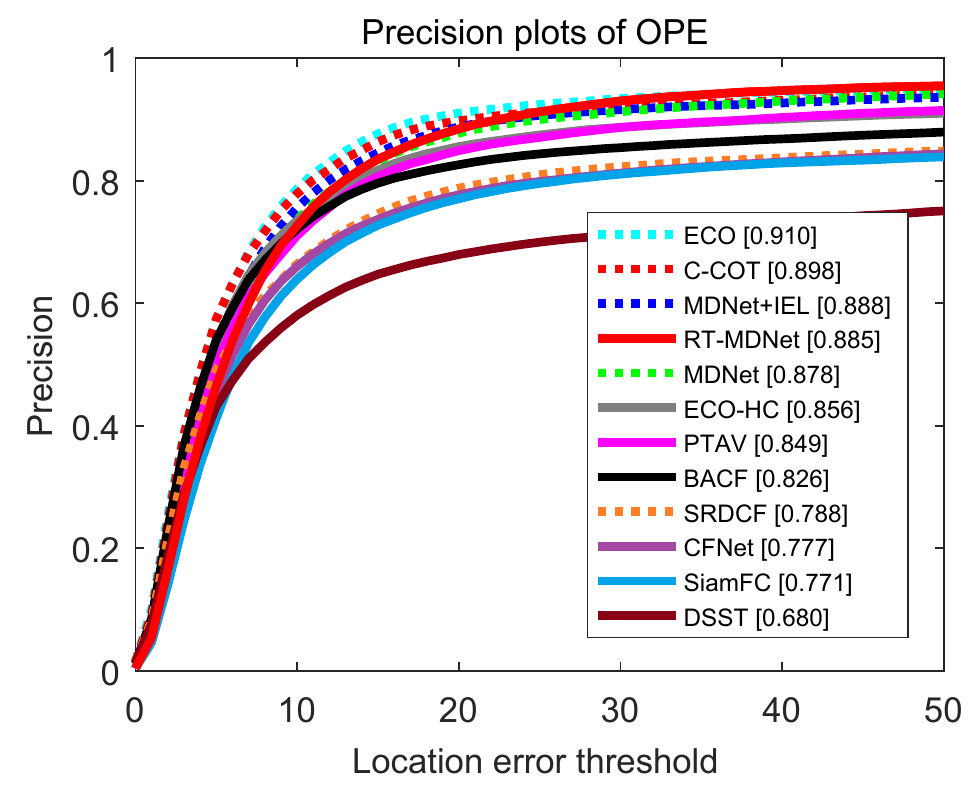}~~~%
    \includegraphics[width=0.49\linewidth]{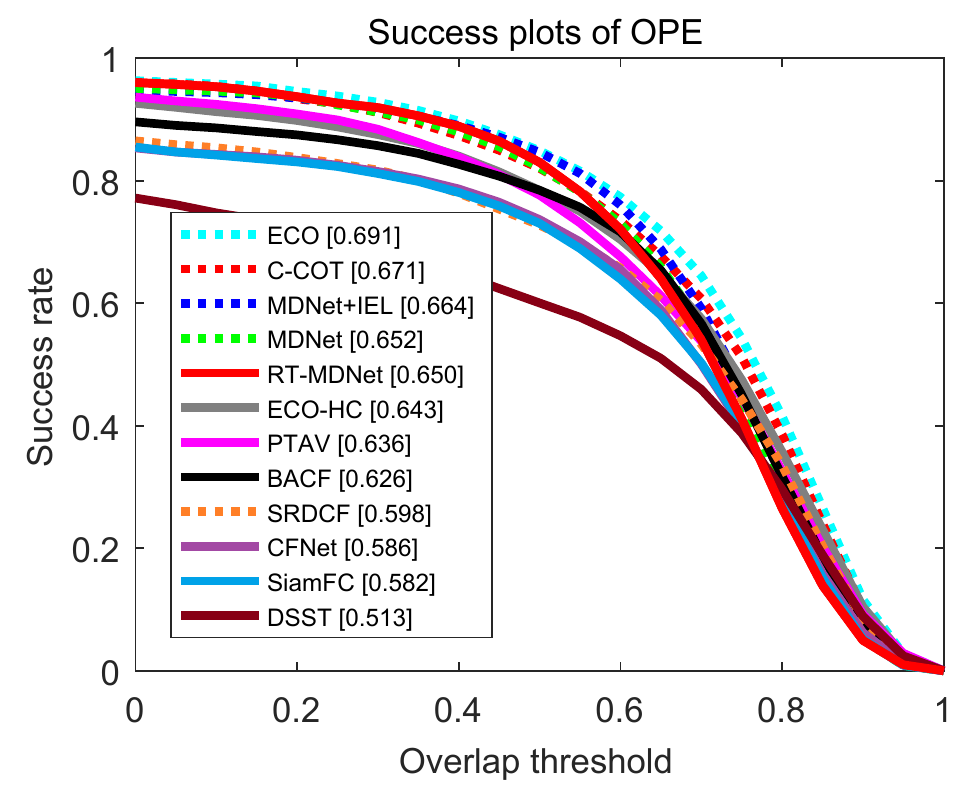}%
    \vspace{-0.3cm} 
\caption{Quantitative results on OTB2015~\cite{otb2015}.}
\label{fig:otb_plot}
\end{figure}

We first analyze our algorithm on OTB2015 dataset~\cite{otb2015}, which consists of 100 fully annotated videos with various challenging attributes.
Fig.~\ref{fig:otb_plot} presents precision and success plots on OTB2015 dataset. 

The results clearly show that real-time MDNet outperforms all the tested real-time trackers significantly in terms of both measures.
It also has competitive accuracy compared to the top-ranked trackers while it is approximately 130, 25, and 8 times faster than C-COT, MDNet, and ECO, respectively.
Our algorithm is slightly less accurate than the competitors when the overlap threshold is larger than 0.8.
It implies that the estimated target bounding boxes given by our tracker are not very tight compared to other state-of-the-art methods; possible reasons are inherent drawback of CNN-based trackers and the limitation of our RoIAlign for target localization at high precision area.

\begin{table}[t!]
\centering
\caption{Quantitative comparisons of real-time trackers on OTB2015}
\label{tab:realtime}
\scalebox{0.9}{
\begin{tabular}{c|ccccccc}
Trackers{}  & {DSST~\cite{dsst}}   & {SiamFC~\cite{siamfc}}    & {CFNet~\cite{CFNET}} & {BACF~\cite{bacf}} & {PTAV~\cite{ptav}} & {ECO-HC~\cite{eco}} & RT-MDNet\\
\hline\hline
Succ (\%) & {51.3}  & {58.2}     & {58.6} & {62.7}& {63.5}& {64.3}  & {65.0} \\
Prec (\%) & {68.0}  & {77.1}     & {77.7} & {82.7}& {84.8}& {85.6}  & {88.5} \\
FPS & {24}     & {86}        & {43}      & {35}   & {25}   & {60}     & {46/52} \\
\hline
\end{tabular}
\vspace{-0.3cm}
}
\end{table}

Table~\ref{tab:realtime} presents overall performance of real-time trackers including our algorithm in terms of AUC for success rate, precision rate at 20 pixel threshold, and speed measured by FPS.
The proposed method outperforms all other real-time trackers by substantial margins in terms of two accuracy measures.
It runs very fast, 46 FPS in average, while speed except the first frame is approximately 52 FPS. 
Note that our tracker needs extra computational cost at the first frame for fine-tuning the network and learning a bounding box regressor.

\begin{figure}[!tb]
\centering
    \includegraphics[width=0.23\linewidth]{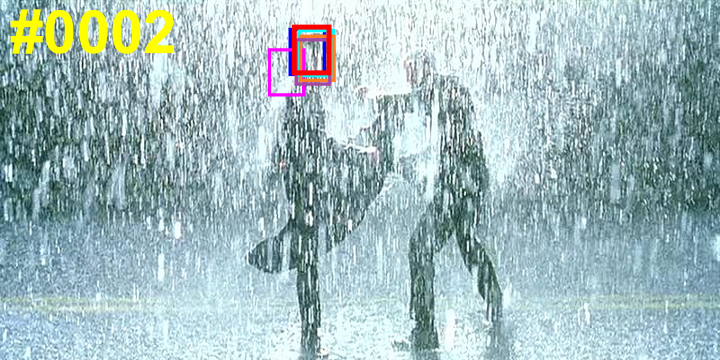}
    \includegraphics[width=0.23\linewidth]{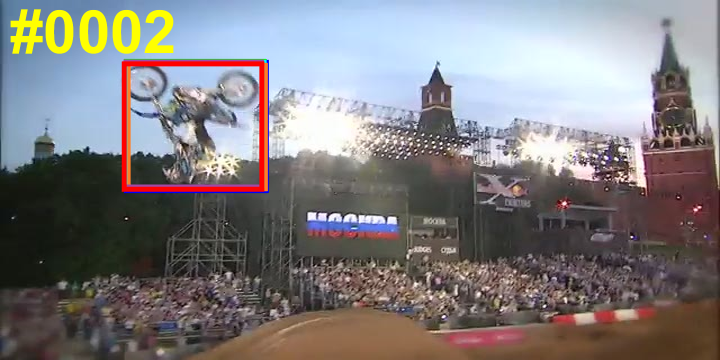}
    \includegraphics[width=0.23\linewidth]{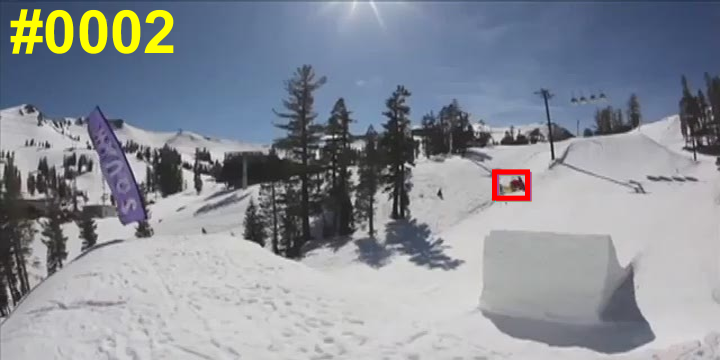}
    \includegraphics[width=0.23\linewidth]{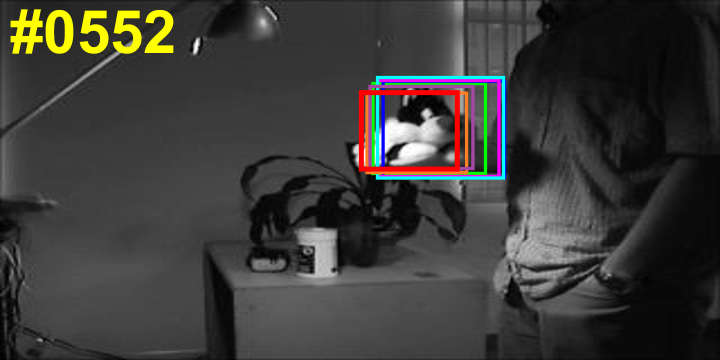}\\
    \includegraphics[width=0.23\linewidth]{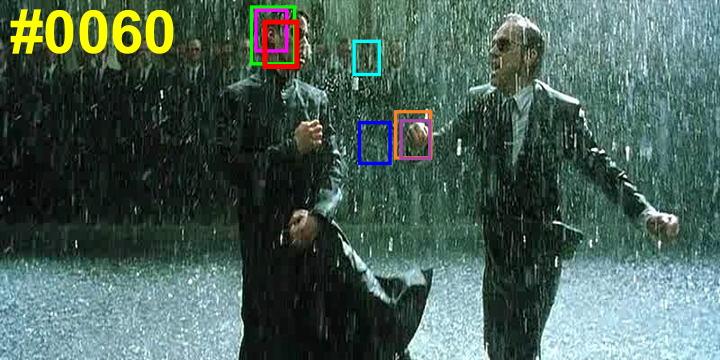}
    \includegraphics[width=0.23\linewidth]{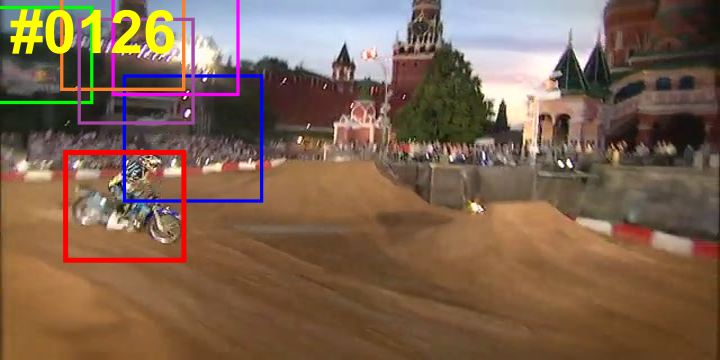}
    \includegraphics[width=0.23\linewidth]{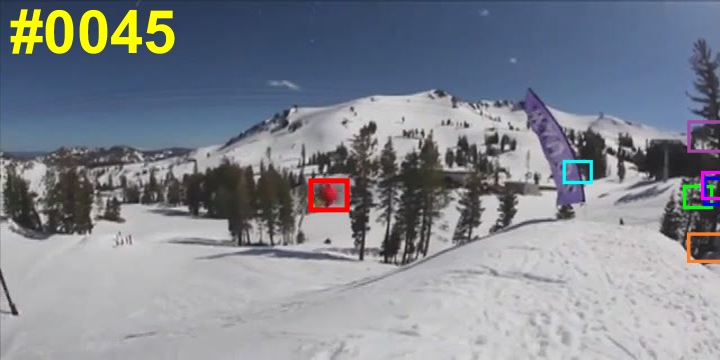}
    \includegraphics[width=0.23\linewidth]{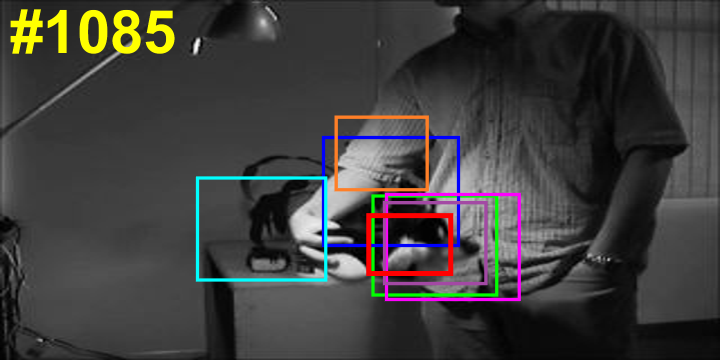}\\
    \includegraphics[width=0.23\linewidth]{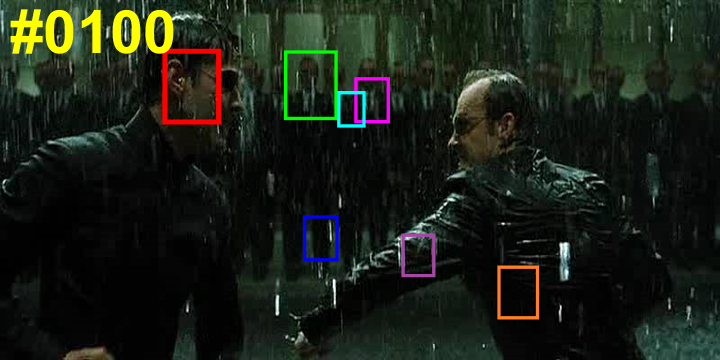}
    \includegraphics[width=0.23\linewidth]{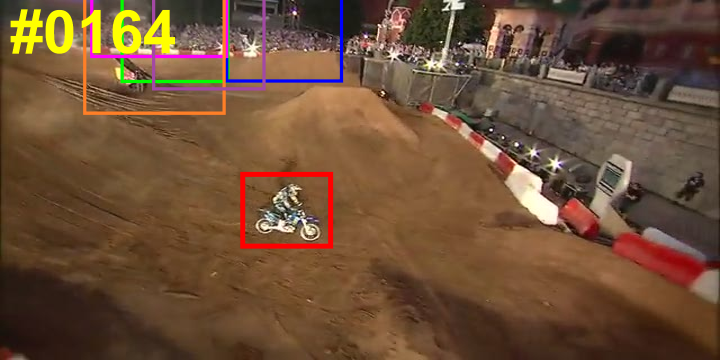}
    \includegraphics[width=0.23\linewidth]{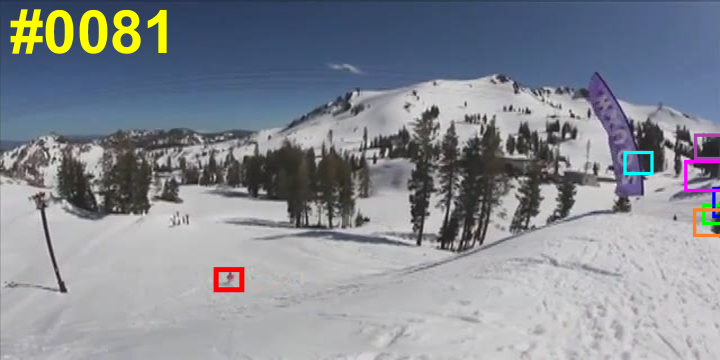}
    \includegraphics[width=0.23\linewidth]{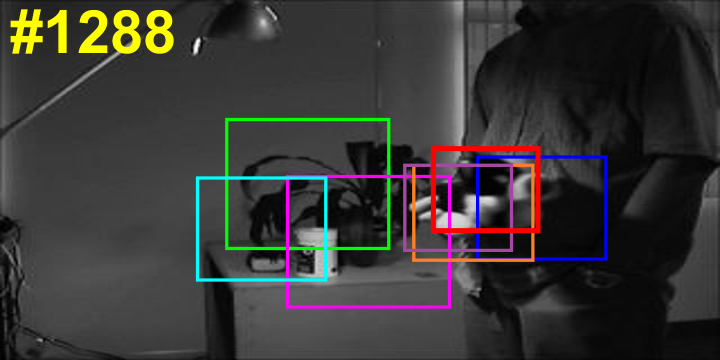}\\
    \includegraphics[width=0.95\linewidth]{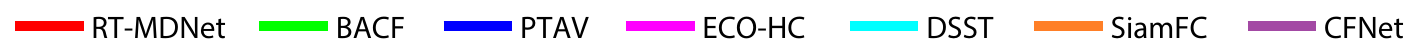}
    \vspace{-0.3cm}
\caption{Qualitative results of the proposed method on several challenging sequences (\textit{Matrix}, \textit{MotorRolling}, \textit{Skiing}, \textit{Sylvester}) in OTB2015 dataset. }

\vspace{-0.3cm}
\label{fig:qualitative_succ}
\end{figure}

\begin{figure}[!tb]
\centering
    \includegraphics[width=0.23\linewidth]{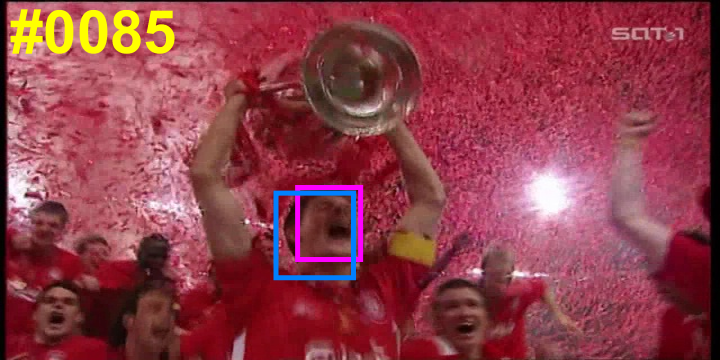}
    \includegraphics[width=0.23\linewidth]{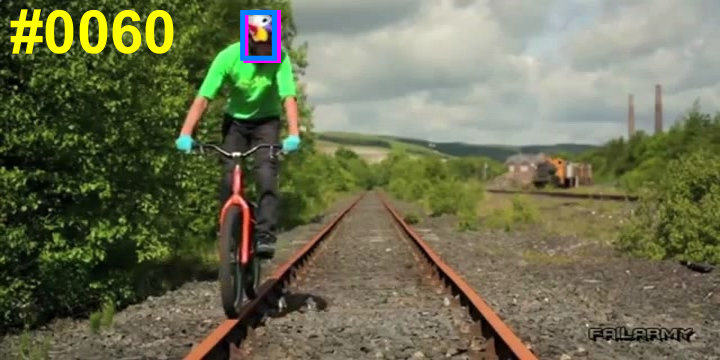}
    \includegraphics[width=0.23\linewidth]{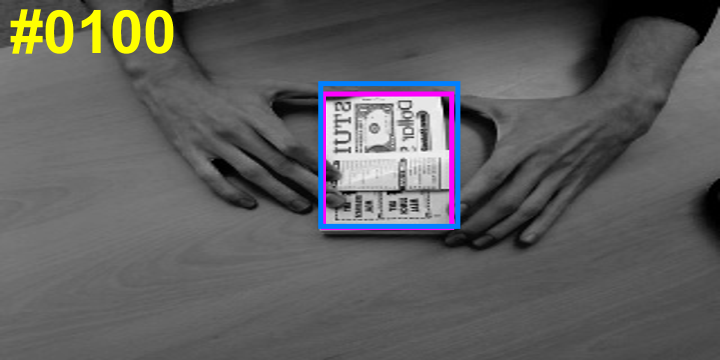}
    \includegraphics[width=0.23\linewidth]{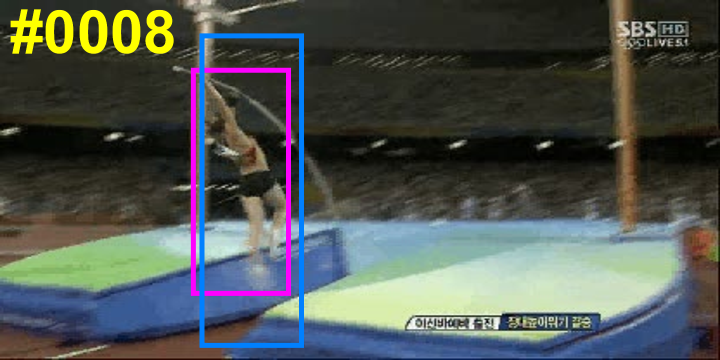}\\
    \includegraphics[width=0.23\linewidth]{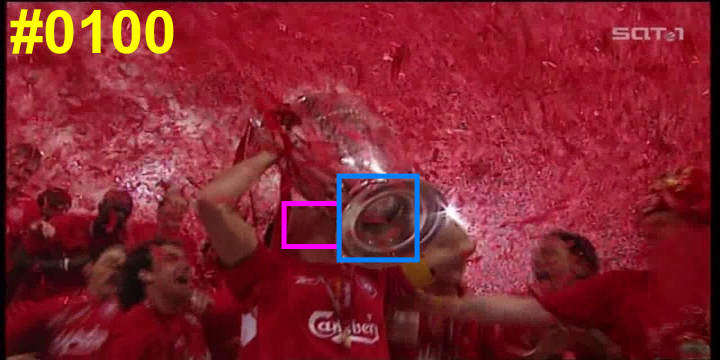}
    \includegraphics[width=0.23\linewidth]{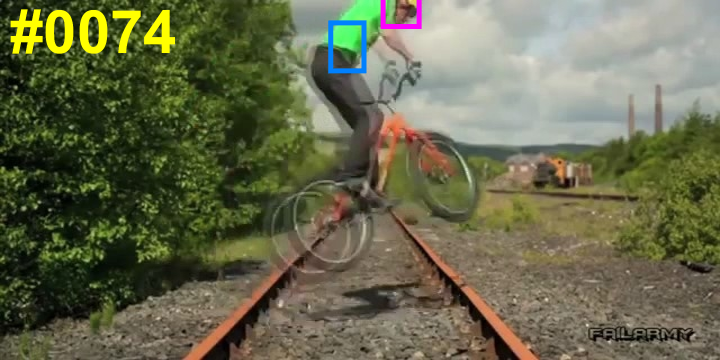}
    \includegraphics[width=0.23\linewidth]{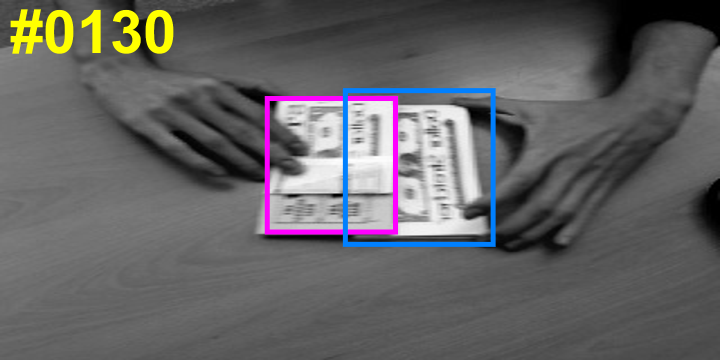}
    \includegraphics[width=0.23\linewidth]{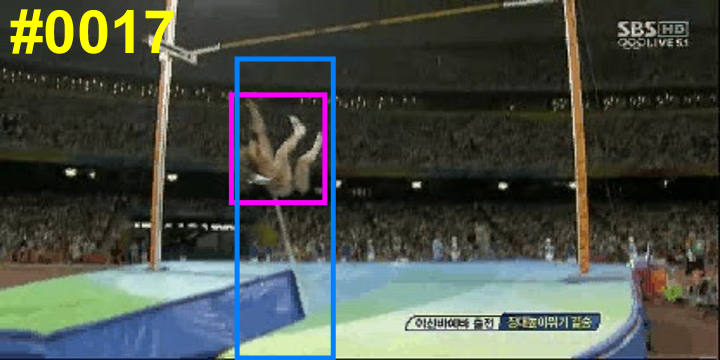}\\
    \includegraphics[width=0.23\linewidth]{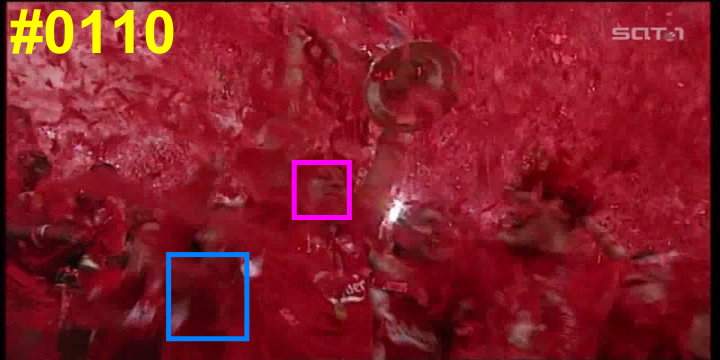}
    \includegraphics[width=0.23\linewidth]{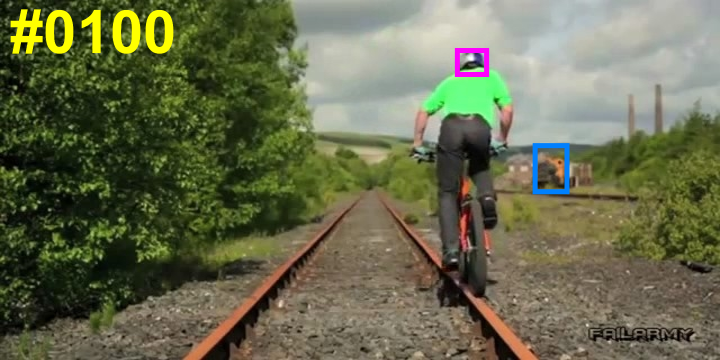} 
    \includegraphics[width=0.23\linewidth]{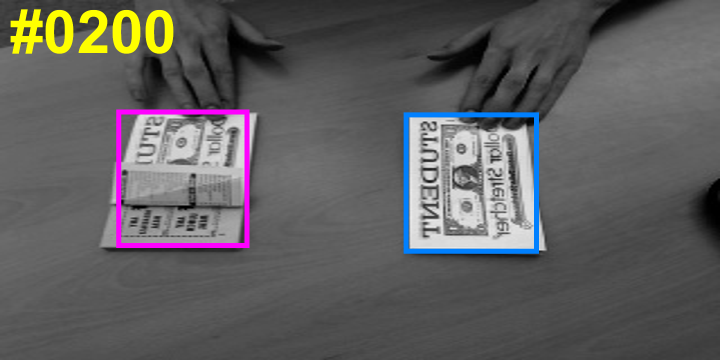}
    \includegraphics[width=0.23\linewidth]{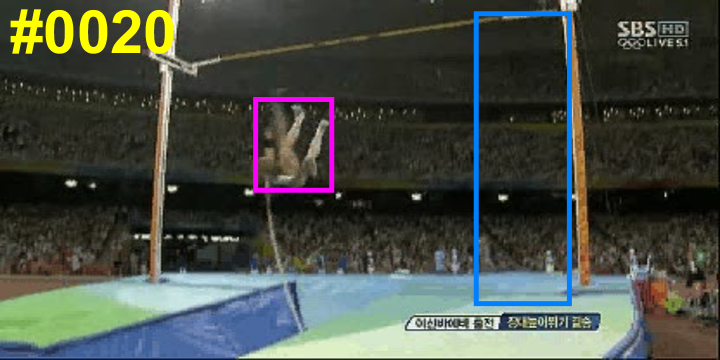}\\
    \vspace{-0.3cm}
\caption{Failure cases of RT-MDNet in \textit{Soccer, Biker, Coupon,} and \textit{Jump} sequence. Magenta and blue bounding boxes denote ground-truths and our results, respectively.}
\vspace{-0.3cm}
\label{fig:qualitative_failure}
\end{figure}

We also illustrate the qualitative results of multiple real-time algorithms on a subset of sequences in Fig.~\ref{fig:qualitative_succ}.
Our approach shows consistently better performance in various challenging scenarios including illumination change, scale variation and background clutter.
Some failure cases are presented in Fig.~\ref{fig:qualitative_failure}.
Our algorithm loses target in \textit{Soccer} sequence due to significant occlusion and in \textit{Biker} sequence due to sudden large motion and out-of-plane rotation. 
Objects with similar appearances make our tracker confused in \textit{Coupon} sequence, and dramatic non-rigid appearance changes in \textit{Jump} cause drift problems.

\subsection{Evaluation on TempleColor}

\begin{figure}[t]
\centering
    \includegraphics[width=0.49\linewidth]{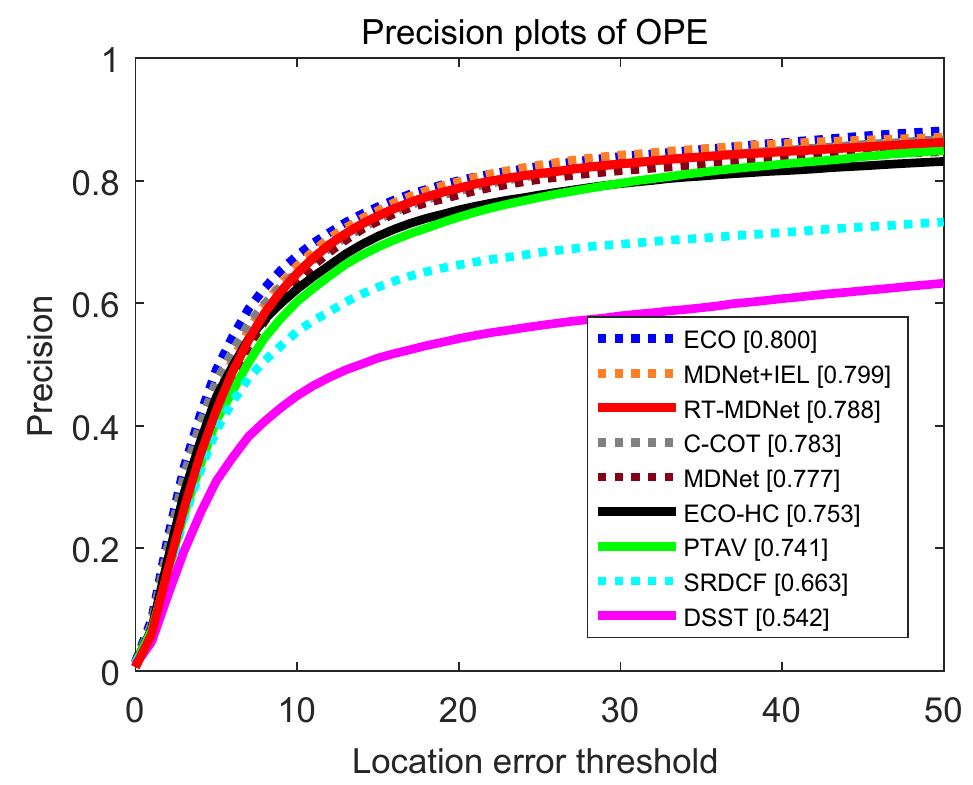}~~~%
    \includegraphics[width=0.49\linewidth]{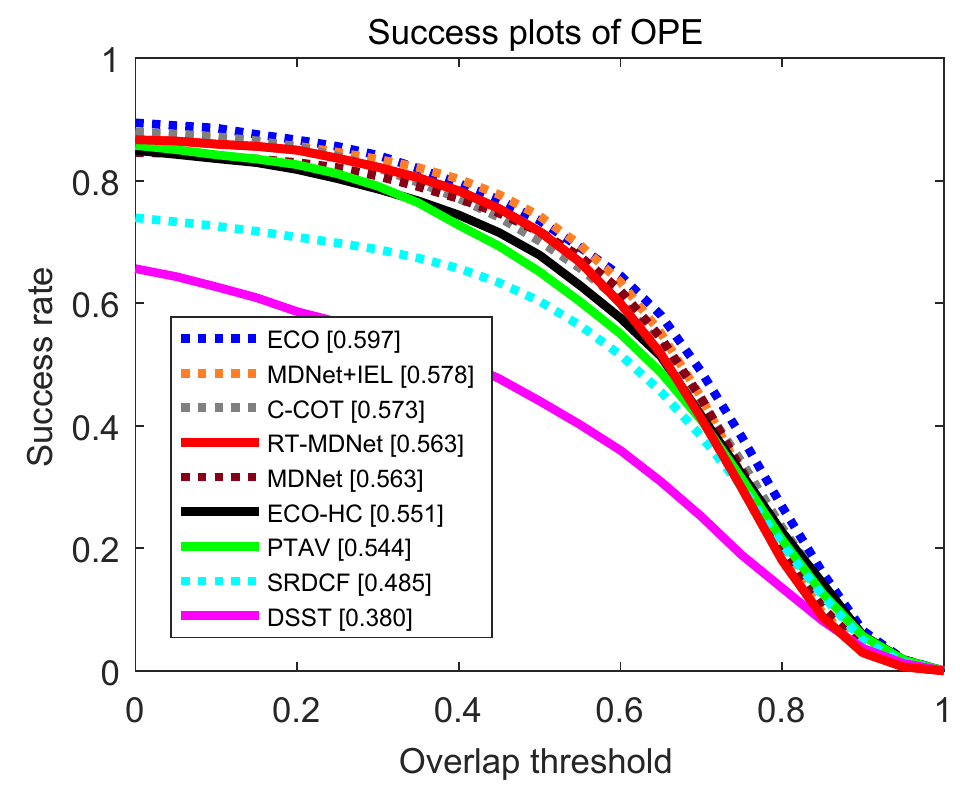}%
    \vspace{-0.3cm}
\caption{Quantitative results on TempleColor~\cite{templecolor}. }
\label{fig:templecolor_plot}
\end{figure}

Fig.~\ref{fig:templecolor_plot} illustrates the precision and success plots on TempleColor dataset~\cite{templecolor}, which is containing 128 color videos while most of sequences are overlapped with OTB2015 dataset~\cite{otb2015}. 
Our method again surpass all real-time trackers\footnote{The AUC score of BACF is reported in their paper by 52.0\%, which is much lower than the score of our tracker.} and has a substantial improvement over ECO-HC.

\subsection{Evaluation on UAV123}

\begin{figure}[t!]
\centering
    \includegraphics[width=0.49\linewidth]{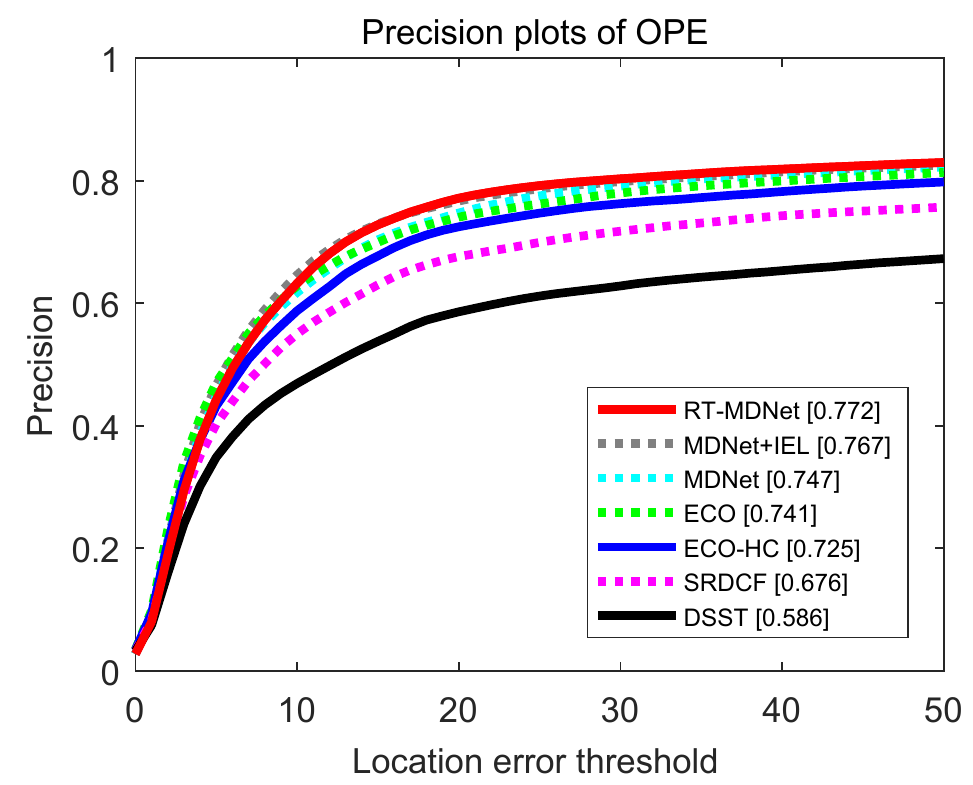}~~~%
    \includegraphics[width=0.49\linewidth]{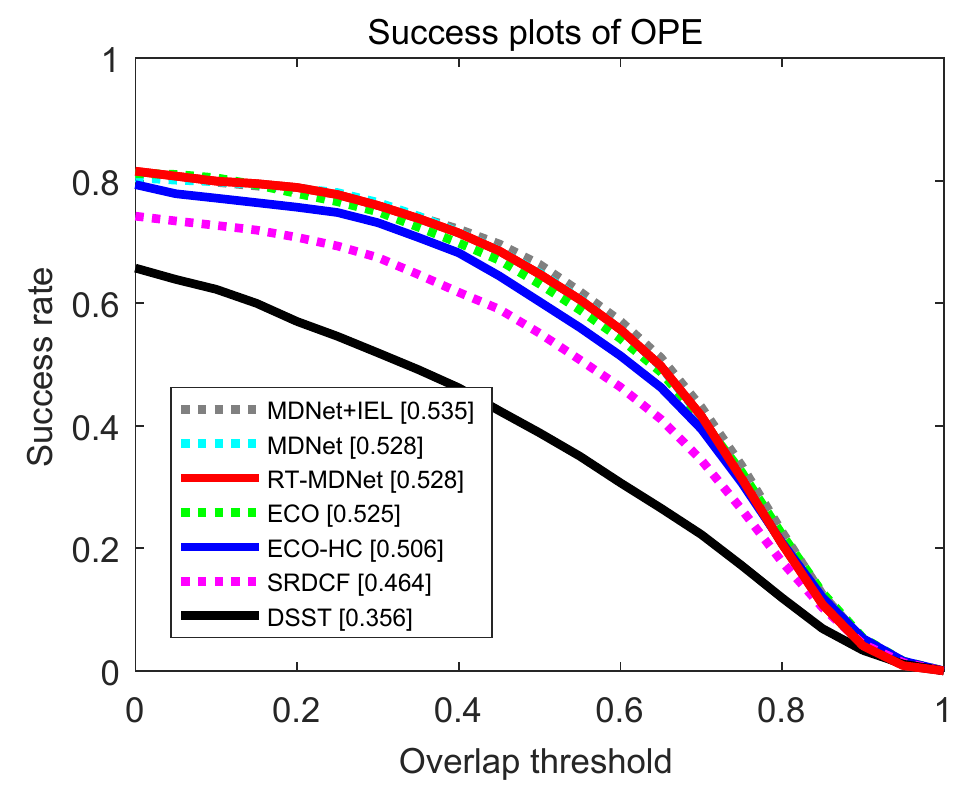}%
    \vspace{-0.3cm}
\caption{Quantitative results on UAV123~\cite{uav123}. }
\label{fig:uav_plot}
\end{figure}

We also evaluate real-time MDNet on the aerial video benchmark, UAV123~\cite{uav123} whose characteristics inherently differ from other datasets such as OTB2015 and TempleColor.
It contains 123 aerial videos with more than 110K frames altogether.
Fig.~\ref{fig:uav_plot} illustrates the precision and success plots of the trackers that have publicly available results on this dataset.
Surprisingly, in the precision rate, our tracker outperforms all the state-of-the-art methods including non-real-time tracker while it is very competitive in the success rate as well.
In particular, our tracker beats ECO, which is a top ranker in OTB2015 and TempleColor, on the both metrics with a approximately 8 times speed-up.
It shows that our algorithm has better generalization capability without parameter tuning to a specific dataset.

\begin{table}[t!]
\centering
\caption{Impacts of different feature extraction methods on accuracy of RT-MDNet}
\label{tab:pooling}
\scalebox{0.9}{
\begin{tabular}{l|ccc|cc}
Pooling Operation & ~align & ~adaRoI & ~denseFM~ & ~Succ (\%) & ~Prec (\%) \\
\hline
\hline
RoIPooling~\cite{fastrcnn} &  &  &   & {35.4} & {53.8}\\
RoIAlign~\cite{maskrcnn} & $\surd$  &  &   & {56.1} & {80.4}\\
Adaptive RoIAlign & $\surd$  & $\surd$ &   & {59.0} & {83.8}\\
RoIAlign with denseFM~ & $\surd$  &   & $\surd$ & {60.7} & {84.3} \\
Improved RoIAlign & $\surd$  & $\surd$  & $\surd$  & {61.9} & {85.3} \\
\hline
\end{tabular}

\vspace{-0.3cm}
}
\end{table}

\subsection{Ablation Study}
\label{subsec:ablation}
We perform several ablation studies on OTB2015~\cite{otb2015} to investigate the effectiveness of individual components in our tracking algorithm.
We first test the impact of the proposed RoIAlign on the quality of our tracking algorithm.
For this experiments, we pretrain our network using {VOT-OTB} dataset, which consist of 58 videos collected from VOT2013~\cite{vot2013}, VOT2014~\cite{vot2014} and VOT2015~\cite{vot2015} excluding the videos in OTB2015.
Table~\ref{tab:pooling} presents several options to extract target representations, which depend on choice between RoIPooling and RoIAlign, use of adaptive RoIAlign layer (adaRoI) and construction of dense feature map (denseFM).
All results consistently support that each component of our improved RoIAlign makes meaningful contribution to tracking performance improvement.

We also investigated two additional versions of our tracking algorithm---one is without bounding box regression (Ours--BBR) and the other is without bounding box regression and instance embedding loss (Ours--BBR--IEL).
Table~\ref{tab:loss} summarizes the results from this internal comparison.
According to our experiment, the proposed multi-task loss (binary classification loss and instance embedding loss)  and bounding box regression are both helpful to improve localization\footnote{As illustrated in Fig.~\ref{fig:otb_plot}, \ref{fig:templecolor_plot}, and \ref{fig:uav_plot}, we verified that applying instance embedding loss to MDNet also improves performances.}.

\begin{table}[t!]
\centering
\caption{Internal comparison results pretrained on ImageNet-Vid dataset.}
\label{tab:loss}
\setlength{\tabcolsep}{10pt}
\scalebox{0.9}{
\begin{tabular}{l|ccc|cc}
Method & $\mathcal{L}_\text{cls}$ & $\mathcal{L}_\text{inst}$ & BBreg & Succ (\%) & Prec (\%)\\
\hline
\hline
Ours--BBR--IEL    &  $\surd$ &         &         & {61.9} & {84.2}\\
Ours--BBR                              &  $\surd$ & $\surd$ &         & {64.1} & {87.7}\\
Ours                                 &  $\surd$ & $\surd$ & $\surd$ & {65.0} & {88.5}\\
\hline
\end{tabular}
\vspace{-0.3cm}
}
\end{table}


\section{Conclusions}
\label{sec:conclusion}
We presented a novel real-time visual tracking algorithm based on a CNN by learning discriminative representations of target in a multi-domain learning framework.
Our algorithm accelerates feature extraction procedure by an improved RoIAlign technique.
We employ a multi-task loss effective to discriminate object instances across domains in the learned embedding space.
The proposed algorithm was evaluated on the public visual tracking benchmark datasets and demonstrated outstanding performance compared to the state-of-the-art techniques, especially real-time trackers.

\subsection*{Acknowledgement}
This research was supported in part by Research Resettlement Fund for the new faculty of Seoul National University  and the IITP grant [2014-0-00059, Development of Predictive Visual Intelligence Technology (DeepView); 2016-0-00563, Research on Adaptive Machine Learning Technology Development for Intelligent Autonomous Digital Companion; 2017-0-01780, The Technology Development for Event Recognition/Relational Reasoning and Learning Knowledge based System for Video Understanding].


\clearpage

\bibliographystyle{splncs}
\bibliography{egbib}
\end{document}